\documentclass{article}

 \usepackage[preprint]{neurips_2026}

\usepackage[utf8]{inputenc} 
\usepackage[T1]{fontenc}    
\usepackage{hyperref}       
\usepackage{url}            
\usepackage{booktabs}       
\usepackage{amsfonts}       
\usepackage{nicefrac}       
\usepackage{microtype}      
\usepackage{xcolor}         
\usepackage{amsmath}
\usepackage{amssymb}
\usepackage{nicefrac}
\usepackage{microtype}
\usepackage{graphicx}
\usepackage{subcaption}
\usepackage{algorithm}
\usepackage{algorithmic}
\usepackage{multirow}
\usepackage{colortbl}
\usepackage{wrapfig}
 
\newcommand{\method}{HELLoRA}
\newcommand{\methodri}{HELLoRI}

\title{HELLoRA: Hot Experts Layer-Level Low-Rank Adaptation for Mixture-of-Experts Models}

\author{%
  Jia Wei \thanks{Corresponding Author. https://weigaa.github.io/.} \\
  Department of Computer Science and Technlogy\\
  Tsinghua University\\
  Beijing 100084, China \\
  \texttt{weijia4473@tsinghua.edu.cn} \\
  \And
  Zhonghao Zhang \\
  School of Computer Science and Technlogy\\
  Xi'an Jiaotong University\\
  \texttt{4125151020@stu.xjtu.edu.cn} \\
  \AND
  Ping Chen \thanks{Corresponding Author.} \\
  The State Key Laboratory of Blockchain and Data Security, Zhejiang University\\
  Hangzhou High-Tech Zone (Binjiang) Institute of Blockchain and Data Security\\
  \texttt{zjuchenping@zju.edu.cn} \\
  \And
  Qianyang Li \\
  School of Computer Science and Technlogy\\
  Xi'an Jiaotong University\\
  \texttt{liqianyang@stu.xjtu.edu.cn} \\
  \And
  Yancheng, Pan \\
  School of Computer Science and Technlogy\\
  Xi'an Jiaotong University\\
  \texttt{3120105301@stu.xjtu.edu.cn} \\
  \And
  Shaoxun Wang \\
  School of Computer Science and Technlogy\\
  Xi'an Jiaotong University\\
  \texttt{shaoxunwang@stu.xjtu.edu.cn} \\
  \And
  Ziyi Qiu \\
  Department of Computer Science and Technlogy\\
  Tsinghua University\\
  \texttt{qiuzy21@mails.tsinghua.edu.cn} \\
  \And
  Longxiang Wang \\
  School of Computer Science and Technlogy\\
  Xi'an Jiaotong University\\
  \texttt{wlx419@xjtu.edu.cn} \\
}

\begin{document}

\maketitle

\begin{abstract}
Low-Rank Adaptation (LoRA) dominates parameter-efficient fine-tuning of
large language models, yet most variants target dense architectures.
Mixture-of-Experts (MoE) models scale parameters at near-constant per-token
compute, and their sparse activation patterns create untapped opportunities
for more efficient adaptation. We propose Hot-Experts Layer-level Low-Rank
Adaptation (\method{}), which attaches LoRA modules only to the most
frequently activated experts at each layer. This simple mechanism reduces
trainable parameters and adapter-induced FLOPs while improving downstream
performance, an effect we attribute to a form of structured regularization
that preserves pretrained expert specialization. To stress-test \method{}
under extreme parameter budgets, we further compose it with LoRI to form
\methodri{}, which freezes the up-projection and sparsifies the
down-projection. Across three MoE backbones, namely OlMoE-1B-7B,
Mixtral-8$\times$7B, and DeepSeekMoE, and three task families covering
mathematical reasoning, code generation, and safety alignment, \method{}
consistently outperforms PEFT baselines. Across three MoE backbones and three task families, \method{} achieves a stronger accuracy--efficiency trade-off than vanilla LoRA. Relative to LoRA, \method{} uses only 16.3\%, 30.1\%, and 23.2\% of the trainable parameters on OlMoE, Mixtral-8$\times$7B, and DeepSeekMoE, respectively, while matching or improving performance on all reported metrics. These results demonstrate that activation-aware adapter placement is an effective and practical route to scaling PEFT for MoE language models.

\end{abstract}

\section{Introduction}
 
Mixture-of-Experts (MoE) models expand parameter capacity while keeping per-token compute nearly constant~\citep{shazeer2017outrageously, yun2024toward}. They match or surpass dense models of comparable compute across diverse domains~\citep{liu2024deepseekv3, jiang2024mixtral, muennighoff2025olmoe}. Despite this efficiency, \emph{adapting} MoE models to downstream tasks remains costly: the full parameter set is large, and naive application of LoRA~\citep{hu2022lora} attaches adapters to all experts, yielding high memory and throughput overhead.
 
\begin{figure}[t]
    \centering
    \begin{subfigure}[b]{0.48\textwidth}
        \includegraphics[trim=100 410 10 135, clip, width=\textwidth]{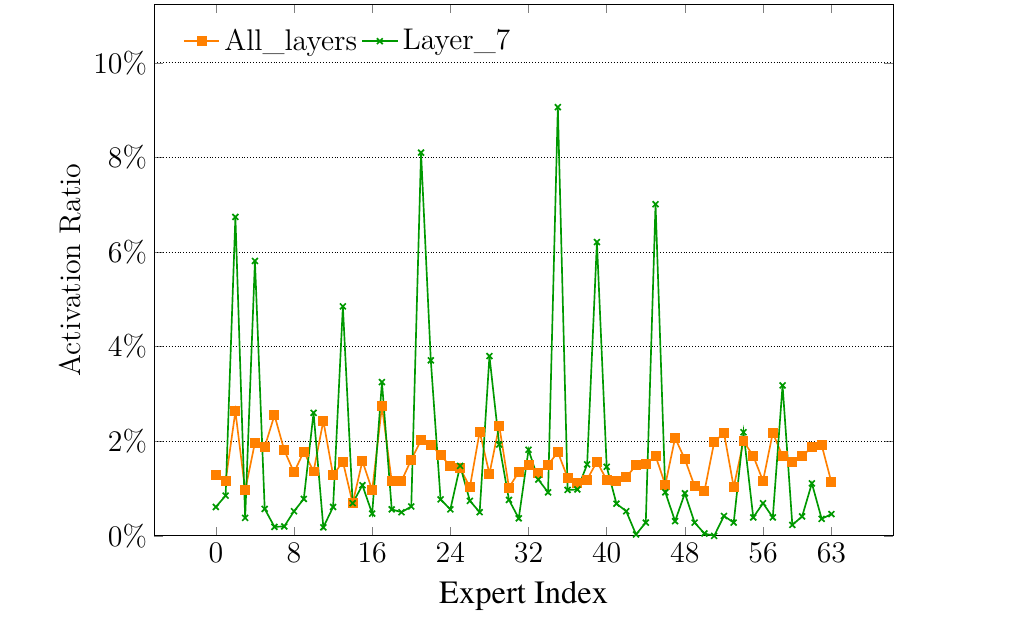}
        \caption{Global vs. Layer-7 activation}
    \end{subfigure}
    \hfill
    \begin{subfigure}[b]{0.48\textwidth}
        \includegraphics[width=\textwidth]{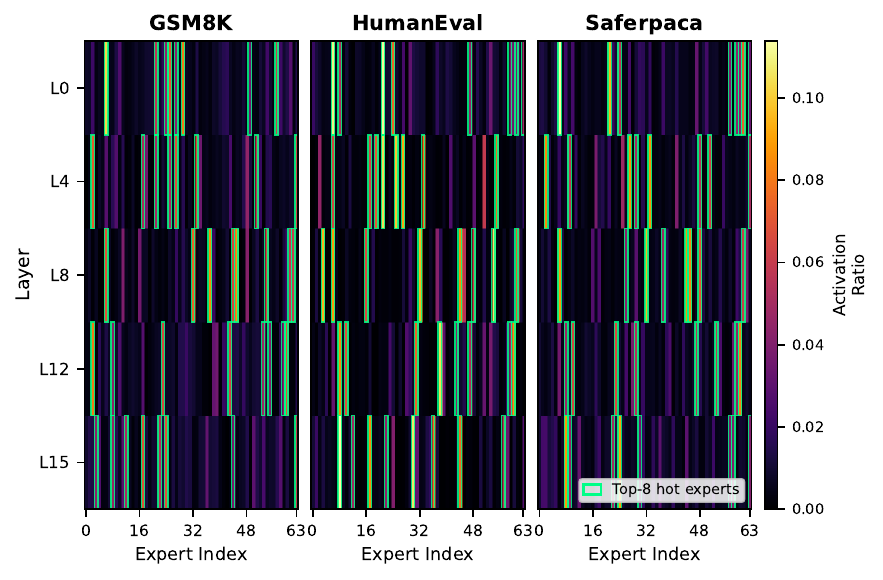}
        \caption{Task-specific activation at different layers}
    \end{subfigure}
    \caption{\textbf{Expert activation patterns in OlMoE.}
    \textbf{(a)}~Across all layers, expert usage appears balanced (orange); within Layer~7 (green), the top-8 experts account for $>$50\% of activations.
    \textbf{(b)}~At the same layer, different tasks activate different expert subsets, confirming that hot experts are both layer-specific and task-specific.}
    \label{fig:motivation}
\end{figure}

A key property of MoE models is \emph{sparse activation}, meaning that only a small subset of experts is activated for each token. Although load-balancing losses encourage more uniform expert utilization during pre-training~\citep{shazeer2017outrageously}, expert activation often becomes highly skewed within each layer on a downstream task, as illustrated in Figure~\ref{fig:motivation}. A small number of \emph{hot} experts account for most token processing, whereas a long tail of \emph{cold} experts is rarely activated. The identity of these hot experts also varies across layers and tasks~\citep{muennighoff2025olmoe}.
 
This observation suggests a simple design principle, namely \textbf{not adapting cold experts}. Attaching LoRA to cold experts wastes parameters on weights that are rarely used for the target task. Worse, it may \emph{hurt} performance by perturbing expert specialization learned during pre-training, thereby injecting gradient noise into parameters whose updates are driven by sparse and unrepresentative token subsets.
 
We instantiate this idea in \textbf{\method{}} (Hot Experts Layer-Level Low Rank Adaptation). \method{} performs a lightweight warm-up on a small subset of the target data to profile expert activation frequencies in each layer. Based on these statistics, it selects the top-$k$ hot experts per layer and inserts LoRA adapters only into those experts. Standard LoRA is applied to attention and gating modules, while cold experts are kept frozen.

\method{} offers two complementary advantages. It improves efficiency by avoiding adapter computation for cold experts in both the forward and backward passes. This is in contrast to masking based methods such as LoRI~\citep{zhang2025lori}, which freeze parameters but still retain forward pass overhead. It also improves regularization by preserving the pretrained specialization of cold experts, thereby reducing cross-task interference. This view is supported by the results in Table~\ref{tab:expert_selection}, where even \emph{random} expert selection slightly outperforms full LoRA. Activation-aware selection yields an additional gain.

To further reduce the parameter budget, we combine \method{} with LoRI~\citep{zhang2025lori} and obtain \textbf{\methodri{}}. This variant freezes the $\mathbf{A}$ matrix and updates only 10\% of the entries in $\mathbf{B}$ within the hot expert adapters.
 
\textbf{Contributions.} We summarize our main contributions as follows.
\begin{enumerate}
    \item We identify a \emph{double sparsity} effect in MoE fine-tuning, 
    where sparse expert activation compounds with the low-rank structure 
    of LoRA updates, and propose \method{}, a layer-level hot-expert 
    adapter placement strategy that exploits this effect.
     \item We evaluate \method{} extensively across three MoE backbones 
    (OlMoE-1B-7B, Mixtral-8$\times$7B, DeepSeekMoE) and three task 
    families (math reasoning, code generation, safety alignment), 
    demonstrating consistent overall gains over both standard PEFT baselines 
    (LoRA, DoRA, LoRI) and MoE-specific adapter methods (LoRAMoE), 
    with thorough ablations on selection strategy, layer choice, 
    warm-up requirements, and hot-expert count.
    \item We show that \method{} simultaneously improves accuracy, 
    parameter efficiency, adapter FLOPs, and wall-clock throughput, a 
    combination that prior masking-based methods do not deliver.
\end{enumerate}
 
\section{Related Work}

\paragraph{LoRA and its variants.}
LoRA~\citep{hu2022lora} injects trainable low-rank factors into frozen linear layers. Subsequent work refines the factorization along several directions. DoRA~\citep{liu2024dora} decomposes weight updates into separate magnitude and direction components. PiSSA~\citep{meng2024pissa} initializes the low-rank factors from the principal singular vectors of the pretrained weight. LoRA-FA~\citep{zhang2023lorafa} freezes the up-projection $\mathbf{A}$ with minimal accuracy loss, roughly halving the number of trainable parameters. LoRI~\citep{zhang2025lori} goes further by additionally sparsifying $\mathbf{B}$, achieving competitive accuracy with an order-of-magnitude fewer updated parameters. These methods all target dense architectures and do not exploit MoE-specific structure.

\paragraph{Mixture-of-Experts models.}
Modern MoE LLMs such as Mixtral~\citep{jiang2024mixtral}, OlMoE~\citep{muennighoff2025olmoe}, and DeepSeekMoE~\citep{liu2024deepseekv3} employ top-$k$ routing with auxiliary load-balancing losses to encourage uniform expert utilization during pre-training. However, when evaluated on specific downstream tasks, per-layer expert usage becomes strongly skewed~\citep{muennighoff2025olmoe}, motivating task-aware adapter placement strategies.

\paragraph{PEFT for MoE models.}
Several lines of work combine LoRA with MoE structure. Adapter-mixture methods such as LoRAMoE~\citep{dou2024loramoe} and 
AdaMoLE~\citep{adamole2024} introduce mixtures over LoRA experts, adding routing mechanisms for the adapters themselves. These methods were designed for dense backbones. In contrast, \method{} exploits the native expert structure of MoE backbones, as our experiments confirm. Dynamic rank-allocation methods~\citep{dynamiclora2024} adjust the LoRA rank across experts during training to redistribute adaptation capacity. These approaches either add architectural complexity or require online rank adjustment. In contrast, \method{} operates at the placement level, deciding which experts receive adapters before training begins through a simple one-shot profiling step. This design makes \method{} orthogonal to both adapter-mixture and rank-allocation strategies, and it can in principle be composed with either family of methods.
 
\section{Method}
 
\subsection{Preliminaries}
 
\paragraph{LoRA.} Given a pretrained weight $\mathbf{W} \in \mathbb{R}^{d_{\text{in}} \times d_{\text{out}}}$, LoRA parameterizes the update as $\Delta = \mathbf{A}\mathbf{B}$ with $\mathbf{A} \in \mathbb{R}^{d_{\text{in}} \times r}$, $\mathbf{B} \in \mathbb{R}^{r \times d_{\text{out}}}$, and $r \ll \min(d_{\text{in}}, d_{\text{out}})$:
\begin{equation}
    \mathbf{h} = \mathbf{x}(\mathbf{W} + \Delta) = \mathbf{x}\mathbf{W} + \mathbf{x}\mathbf{A}\mathbf{B}.
    \label{eq:lora}
\end{equation}
 
\paragraph{MoE load-balancing.} An MoE layer contains $N_E$ experts $\{E_i\}_{i=1}^{N_E}$ with top-$k$ routing. The auxiliary load-balancing loss~\citep{shazeer2017outrageously}
\begin{equation}
    \mathcal{L}_{\text{LB}} = N_E \cdot \sum_{i=1}^{N_E} f_i \cdot P_i
    \label{eq:lb}
\end{equation}
encourages uniform expert usage during pre-training. Here $f_i$ is the token fraction assigned to expert $E_i$ and $P_i$ is the mean routing probability. Crucially, widely-used implementations (OlMoE, Mixtral) compute $\mathcal{L}_{\text{LB}}$ over concatenated router logits across all layers, yielding approximately balanced \emph{global} usage but potentially imbalanced \emph{per-layer} usage on specific tasks.
 
\subsection{\method{}: Hot-Experts Layer-Level Adapter Placement}
 
\begin{figure}[t]
    \centering
     \includegraphics[width=0.9\textwidth]{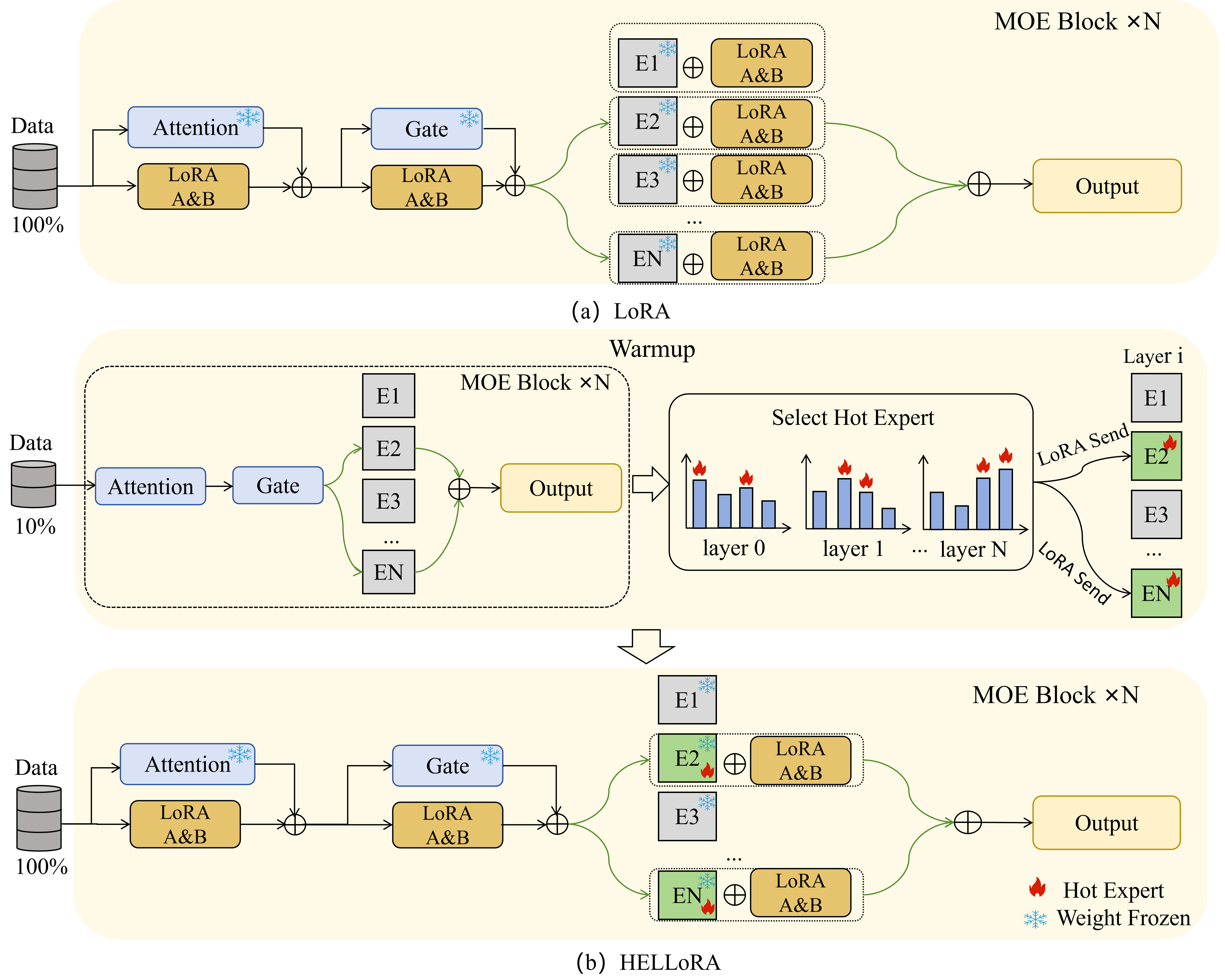}
    \caption{\textbf{Comparison of LoRA and \method{}.}
    Standard LoRA attaches adapters to all experts in every MoE layer (top).
    \method{} attaches adapters only to hot experts identified by the warm-up profiling stage, keeping cold experts frozen (bottom).
    Attention and gating components receive LoRA adapters in both settings.}
    \label{fig:method_overview}
\end{figure}

As shown in Figure \ref{fig:method_overview}, \method{} consists of two stages: (1) a lightweight profiling stage to identify hot experts, and (2) standard LoRA fine-tuning with adapters placed only on hot experts.
 
\paragraph{Stage 1: Layer-level hot-expert profiling.}
We run a short LoRA warm-up on a random $p\%$ sample of the target training data (default $p = 10$). During this warm-up, we log the activation count $c_i^{(\ell)}$ of each expert $i$ at each MoE layer $\ell$. We then sort experts by $c_i^{(\ell)}$ within each layer and select the top-$k$ as \emph{hot experts}. The output is a per-layer hot-expert list $\mathcal{H}^{(\ell)} = \{i : c_i^{(\ell)} \text{ is among the top-}k\}$. Algorithm~\ref{alg:hellora} summarizes the full procedure.
 
\paragraph{Stage 2: Selective adapter attachment.}
We attach LoRA modules to all attention projections, the gating network, and only the hot experts in $\mathcal{H}^{(\ell)}$ at each MoE layer $\ell$. Cold experts remain fully frozen. The base model is frozen throughout and only the adapter parameters are trained.
 
\begin{algorithm}[t]
\caption{\method{}: Hot-Experts Layer-Level Low-Rank Adaptation}
\label{alg:hellora}
\begin{algorithmic}[1]
\REQUIRE Pretrained MoE model $\mathcal{M}$, target dataset $\mathcal{D}$, warm-up fraction $p$, hot-expert count $k$
\STATE Sample $\mathcal{D}_{\text{warm}} \sim \mathcal{D}$ with $|\mathcal{D}_{\text{warm}}| = p\% \cdot |\mathcal{D}|$
\STATE Run LoRA warm-up on $\mathcal{D}_{\text{warm}}$; log per-layer expert activation counts $\{c_i^{(\ell)}\}$
\FOR{each MoE layer $\ell$}
    \STATE $\mathcal{H}^{(\ell)} \leftarrow \text{top-}k$ experts by $c_i^{(\ell)}$
\ENDFOR
\STATE Attach LoRA to: attention projections, gating networks, and experts in $\{\mathcal{H}^{(\ell)}\}_\ell$
\STATE Fine-tune on full dataset $\mathcal{D}$ (only adapter parameters are trainable)
\RETURN Adapted model
\end{algorithmic}
\end{algorithm}
 
\paragraph{Why hot experts?} We provide two complementary perspectives on why \method{} improves over full LoRA.

\emph{Efficiency.} Removing adapters from cold experts eliminates their adapter forward and backward passes entirely. Unlike masking methods such as LoRI, which freeze parameters but retain them in the computational graph, \method{} reduces actual FLOPs and memory traffic. The adapter FLOPs scale with $|\mathcal{H}^{(\ell)}|$ rather than $N_E$.

\emph{Regularization.} Pre-trained MoE models embed task-relevant knowledge in expert specialization. Adapting cold experts that are rarely activated for the target task injects noisy gradients computed from sparse, unrepresentative token subsets. Freezing cold experts preserves their pretrained function, acting as a form of structured regularization. Empirically, even random expert selection, which also restricts scope, slightly improves over full LoRA, while activation-aware selection yields substantially larger gains (Section~\ref{sec:ablation_expert}).
 
\subsection{\methodri{}: Extreme Parameter Budgets}
 
To explore the lower bound of trainable parameters, we compose \method{} with LoRI~\citep{zhang2025lori}. \methodri{} freezes $\mathbf{A}$ (initialized randomly and fixed) and updates only the top 10\% of $\mathbf{B}$'s entries by absolute value:
\begin{equation}
    \mathbf{h} = \mathbf{x}\mathbf{W} + \mathbf{x}\mathbf{A}(\mathbf{B} \odot \mathbf{M}),
    \label{eq:hellori}
\end{equation}
where $\mathbf{M}$ is a binary mask selecting the top-10\% entries. This yields $\sim$0.03\% trainable parameters relative to full fine-tuning.
 
\section{Experiments}
 
\subsection{Setup}
 
\paragraph{Models.} We evaluate on three MoE backbones spanning different architectures and scales:
\begin{itemize}
    \item \textbf{OlMoE-1B-7B}~\citep{muennighoff2025olmoe}: 7B total parameters, 1B activated per step, 16 layers, 64 experts/layer, top-8 routing.
    \item \textbf{Mixtral-8$\times$7B}~\citep{jiang2024mixtral}: 47B total parameters, 13B activated per step, 32 layers, 8 experts/layer, top-2 routing.
    \item \textbf{DeepSeekMoE}~\citep{liu2024deepseekv3}: A hybrid architecture combining dense layers, shared experts, and 64 routed experts per MoE layer
with top-6 routing.
\end{itemize}
 
\paragraph{Tasks.}
(1)~\textbf{Math reasoning}: fine-tune on GSM8K~\citep{cobbe2021training} train, report test accuracy.
(2)~\textbf{Code generation}: fine-tune on CodeAlpaca~\citep{chaudhary2023code}, evaluate on HumanEval~\citep{chen2021evaluating} (pass@1/5/10).
(3)~\textbf{Safety alignment}: fine-tune on Saferpaca~\citep{bianchi2023safety}, evaluate refusal rate on HEx-PHI~\citep{qi2024finetuning}.
 
\textbf{Baselines.} We compare \method{} against full fine-tuning (FFT), 
LoRA~\citep{hu2022lora}, DoRA~\citep{liu2024dora}, LoRI-D and 
LoRI-S~\citep{zhang2025lori}, and LoRAMoE~\citep{dou2024loramoe}. 
A rank-matched variant of LoRAMoE is reported in 
Appendix~\ref{app:loramoe_rank32}.

\textbf{Training details.} All experiments use FSDP~\citep{zhao2023pytorch} 
on two NVIDIA A100 GPUs. For \method{}, LoRA, DoRA, and LoRI variants, 
we set LoRA rank $r = 32$ and scaling factor $\alpha = 64$. For LoRAMoE, 
we follow the original configuration with rank $r = 4$ and $N = 6$ LoRA 
experts per FFN. The number of hot experts $k$ is set to match the 
routing budget of each backbone, yielding $k = 8$ for OlMoE, $k = 2$ 
for Mixtral, and $k = 12$ for DeepSeekMoE. We discuss the rationale for 
$k = 12$ on DeepSeekMoE in Section~\ref{sec:k_ablation}. The warm-up 
uses 10\% of the training data, and results are averaged over 10 random 
seeds. Full hyperparameters are in Appendix~\ref{app:hyperparams}.
 
\begin{figure}[t]
    \centering
    \includegraphics[width=\textwidth]{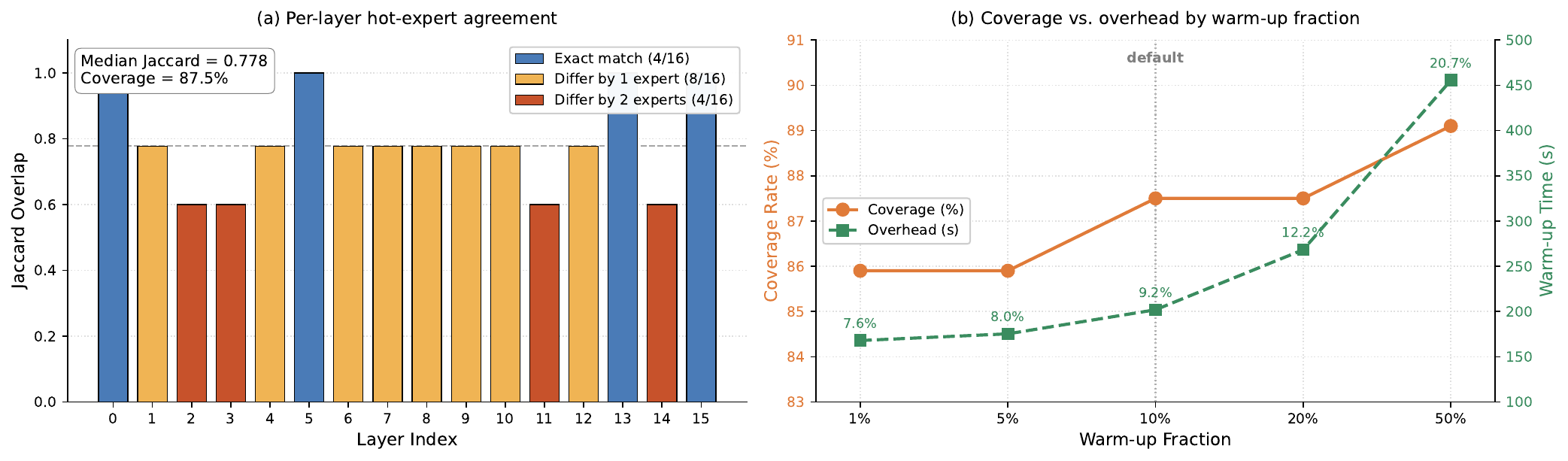}
    \caption{\textbf{Warm-up stability analysis.} (a) Jaccard overlap between hot experts identified using 10\% of the data and those identified using the full dataset. 4 out of 16 layers match exactly, 8 layers differ by one expert, and 4 layers differ by two experts. (b) Coverage of full-data hot experts and wall-clock overhead as the warm-up fraction increases. Using 10\% of the data, our default setting, recovers 87.5\% of the hot experts at only 9.2\% additional training cost.}
    \label{fig:warmup_validation}
\end{figure}
 
\subsection{Main Results}
\label{sec:main_results}
 
\textbf{OlMoE results.} Table~\ref{tab:main_olmoe} reports results across 
three tasks. \method{} outperforms all baselines on GSM8K with an accuracy 
of $29.49_{\pm 0.32}$, compared with $26.37_{\pm 1.07}$ for LoRA and 
$26.88_{\pm 1.28}$ for LoRAMoE. It also leads on HumanEval pass@1 and 
pass@5 and achieves the highest safety score of 99.06. Notably, \method{} 
uses only 0.7\% trainable parameters, roughly six times fewer than LoRA 
and five times fewer than LoRAMoE. The gap over LoRAMoE 
(Appendix~\ref{app:loramoe_rank32}) reflects a structural mismatch 
between adapter-level routing and the native routing of MoE backbones.
 
\begin{table}[t]
\centering
\caption{Accuracy comparison on OlMoE-1B-7B. Bold indicates the best value per column. \method{} achieves the best accuracy with $\sim$6$\times$ fewer trainable parameters than LoRA. LoRAMoE uses rank $r = 4$ with $N = 6$ experts per FFN, following the original configuration. Rank-matched results are reported in Appendix~\ref{app:loramoe_rank32}. Results are mean $\pm$ standard deviation over 10 random seeds.}
\label{tab:main_olmoe}
\begin{tabular}{lccccc}
\toprule
Method & Params & GSM8K & HumanEval (@1 / @5 / @10) & HEx-PHI \\
\midrule
FFT        & 100\%   & $27.64_{\pm 0.23}$ & $16.40_{\pm 0.17}$ / $21.04_{\pm 0.21}$ / $22.38_{\pm 0.33}$ & $97.32_{\pm 1.02}$ \\
LoRA       & 4.300\% & $26.37_{\pm 1.07}$ & $16.52_{\pm 0.21}$ / $21.80_{\pm 0.30}$ / $\mathbf{24.03_{\pm 0.72}}$ & $98.75_{\pm 0.39}$ \\
DoRA       & 4.350\% & $26.82_{\pm 0.83}$ & $16.72_{\pm 0.11}$ / $21.93_{\pm 0.14}$ / $23.68_{\pm 0.36}$ & $98.86_{\pm 0.11}$ \\
LoRAMoE       & 3.594\% & $26.88_{\pm 1.28}$ & $15.64_{\pm 0.12}$ / $19.69_{\pm 0.31}$ / $20.64_{\pm 0.54}$ & $98.13_{\pm 0.02}$ \\
LoRI-D     & 1.900\% & $28.22_{\pm 0.73}$ & $16.98_{\pm 0.12}$ / $21.85_{\pm 0.19}$ / $23.75_{\pm 0.07}$ & $96.56_{\pm 1.31}$ \\
LoRI-S     & 0.190\% & $25.49_{\pm 2.33}$ & $16.46_{\pm 0.25}$ / $20.22_{\pm 0.33}$ / $22.37_{\pm 0.51}$ & $98.44_{\pm 0.44}$ \\
\midrule
HELLoRA    & 0.678\% & $\mathbf{29.49_{\pm 0.32}}$ & $\mathbf{17.87_{\pm 0.11}}$ / $\mathbf{22.04_{\pm 0.06}}$ / $23.82_{\pm 0.32}$ & $\mathbf{99.06_{\pm 0.04}}$ \\
HELLoRI-D  & 0.301\% & $28.42_{\pm 0.61}$ & $16.89_{\pm 0.13}$ / $20.89_{\pm 0.47}$ / $22.21_{\pm 0.49}$ & $98.75_{\pm 0.32}$ \\
HELLoRI-S  & 0.030\% & $25.10_{\pm 1.71}$ & $16.22_{\pm 0.33}$ / $20.27_{\pm 0.55}$ / $21.76_{\pm 0.77}$ & $97.50_{\pm 0.71}$ \\
\bottomrule
\end{tabular}
\end{table}
 

\paragraph{Cross-backbone validation.} Table~\ref{tab:cross_backbone} 
extends our analysis to Mixtral-8$\times$7B and DeepSeekMoE across all three task families. \method{} consistently outperforms LoRA on all nine backbone-task combinations while using only 16 to 30 percent of LoRA's trainable parameters. On DeepSeekMoE, where shared experts are always active and receive adapters under both methods, \method{} matches LoRA's accuracy at roughly four times fewer trainable parameters. These results confirm that hot-expert placement generalizes across MoE architectures with diverse routing designs (top-8 in OlMoE, top-2 in Mixtral, and shared-plus-routed in DeepSeekMoE) and across task types ranging from mathematical reasoning to code generation and safety alignment.


\begin{table}[t]
\centering
\small
\setlength{\tabcolsep}{4pt}
\caption{Cross-backbone comparison of LoRA and \method{} on three task 
families. \method{} matches or outperforms LoRA on all backbone-task 
combinations while using only 16\% to 30\% of the trainable parameters 
used by LoRA. We set $k = 8$ for OlMoE, $k = 2$ for Mixtral, and $k = 12$ 
for DeepSeekMoE. Full results including LoRI and \methodri{} variants 
are provided in Appendix~\ref{app:cross_backbone_full}. Results are 
mean $\pm$ standard deviation over 10 random seeds.}
\label{tab:cross_backbone}
\begin{tabular}{llcccc}
\toprule
Model & Method & Params & GSM8K & HumanEval (@1 / @5 / @10) & HEx-PHI \\
\midrule
\multirow{2}{*}{OlMoE-1B-7B}    
    & LoRA      & 4.300\% & $26.37_{\pm 1.07}$ & $16.52_{\pm 0.21}$ / $21.80_{\pm 0.30}$ / $24.03_{\pm 0.72}$ & $98.75_{\pm 0.39}$ \\
    & \method{} & 0.678\% & $\mathbf{29.49_{\pm 0.32}}$ & $\mathbf{17.87_{\pm 0.11}}$ / $\mathbf{22.04_{\pm 0.06}}$ / $23.82_{\pm 0.32}$ & $\mathbf{99.06_{\pm 0.04}}$ \\
\midrule
\multirow{2}{*}{Mixtral-8$\times$7B}    
    & LoRA      & 1.030\% & $67.73_{\pm \text{2.01}}$ & $\text{39.99}_{\pm \text{2.46}}$ / $\text{49.47}_{\pm \text{2.68}}$ / $\text{52.72}_{\pm \text{2.10}}$ & $93.94_{\pm \text{0.59}}$ \\
    & \method{} & 0.310\% & $\mathbf{68.28_{\pm \text{1.19}}}$ & $\mathbf{44.89_{\pm 2.02}}$ / $\mathbf{53.97_{\pm 1.84}}$ / $\mathbf{57.41_{\pm 2.23}}$ & $\mathbf{96.67}_{\pm 0.23}$ \\
\midrule
\multirow{2}{*}{DeepSeekMoE}    
    & LoRA      & 3.545\% & $35.15_{\pm \text{0.86}}$ & $\text{27.92}_{\pm \text{1.25}}$ / $\text{34.98}_{\pm \text{1.76}}$ / $\text{38.52}_{\pm \text{0.82}}$ & $\text{97.87}_{\pm \text{1.22}}$ \\
    & \method{} & 0.823\% & $\mathbf{35.54_{\pm \text{0.66}}}$ & $\mathbf{31.15}_{\pm 1.39}$ / $\mathbf{36.65}_{\pm 2.03}$ / $\mathbf{39.19}_{\pm 1.01}$ & $\mathbf{98.49}_{\pm 0.83}$ \\
\bottomrule
\end{tabular}
\end{table}
 
\subsection{Efficiency Analysis}
\label{sec:efficiency}
 
\begin{figure}[t]
    \centering
    \includegraphics[width=0.75\textwidth]{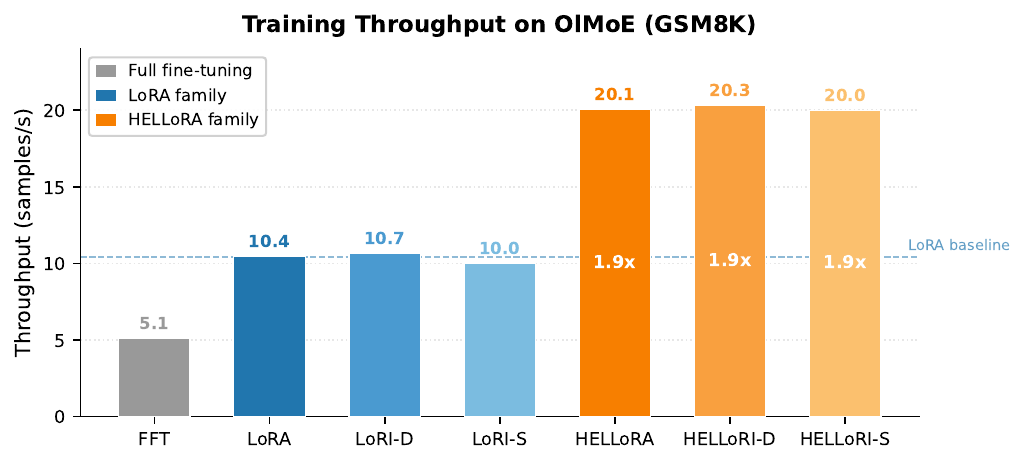}
    \caption{\textbf{Training throughput on OlMoE (GSM8K).}
    \method{} and its variants achieve approximately 1.9$\times$ the throughput
    of LoRA by removing adapter kernels from cold experts entirely.
    Unlike masking-based methods (LoRI), which retain all adapter parameters
    in the computational graph, \method{} reduces actual FLOPs and memory traffic,
    yielding a genuine wall-clock speedup.}
    \label{fig:throughput}
\end{figure}

\paragraph{Training throughput.} Figure~\ref{fig:throughput} shows end-to-end throughput on OlMoE. \method{} and \methodri{} achieve $\sim$1.9$\times$ the throughput of LoRA/LoRI variants and $\sim$3.9$\times$ that of FFT. The warm-up stage adds $\sim$9\% overhead (201.86s out of 2204.48s total training time).
 
\paragraph{Adapter-induced FLOPs.} Table~\ref{tab:flops} reports the
incremental FLOPs introduced by the PEFT modules. Backbone FLOPs are
identical across methods and therefore omitted. \method{} reduces adapter
FLOPs by 33 to 39\% across all three backbones. The FLOPs reduction is
smaller than the parameter reduction because in MoE models only activated
experts execute their adapters, so the per-step compute savings are bounded
by the activation ratio rather than the total expert count.
 
\begin{table}[t]
\centering
\caption{Adapter-induced FLOPs (arbitrary units). \method{} reduces adapter compute by 33--39\% across backbones. Backbone FLOPs (identical across methods) are omitted.}
\label{tab:flops}
\small
\begin{tabular}{@{}llcc@{}}
\toprule
\textbf{Model} & \textbf{Method} & \textbf{Forward FLOPs} & \textbf{Train FLOPs} \\
\midrule
\multirow{2}{*}{OlMoE} & LoRA & 94.44 & 283.31 \\
& \method{} & 57.89 (\textcolor{teal}{--38.7\%}) & 173.64 (\textcolor{teal}{--38.7\%}) \\
\midrule
\multirow{2}{*}{Mixtral} & LoRA & 289.42 & 868.27 \\
& \method{} & 195.66 (\textcolor{teal}{--32.4\%}) & 586.97 (\textcolor{teal}{--32.4\%}) \\
\midrule
\multirow{2}{*}{DeepSeekMoE} & LoRA & 170.77 & 512.31 \\
& \method{} & 113.06 (\textcolor{teal}{--33.8\%}) & 339.18 (\textcolor{teal}{--33.8\%}) \\
\bottomrule
\end{tabular}
\end{table}
 
\paragraph{Comparison with masking methods.}
LoRI reduces trainable parameters via masking but does not reduce compute: the mask introduces additional reads/writes and prevents kernel fusion. LoRI-S achieves similar throughput to LoRA (Table~\ref{tab:layer_throughput}). In contrast, \method{} removes adapter kernels entirely from cold experts, yielding a genuine wall-clock speedup. \method{} is the first approach to simultaneously reduce parameters \emph{and} improve throughput for MoE PEFT.

 
\subsection{Ablation Studies}
\label{sec:ablations}
 
\subsubsection{Expert Selection Strategy}
\label{sec:ablation_expert}
 
We compare four expert selection strategies on GSM8K (Table~\ref{tab:expert_selection}): (1) \emph{layer-level hot} (our default), (2) \emph{model-level hot} (global top-$k$ applied uniformly), (3) \emph{cold experts} (bottom-$k$ per layer), and (4) \emph{random} (uniform random $k$ per layer, averaged over 5 seeds). We observe a clear ordering of strategies. Random selection itself slightly outperforms full LoRA, indicating that scope restriction provides regularization. Layer-level hot selection further exceeds random by 2.4 points, demonstrating that adapter \emph{placement} matters. Cold-expert placement underperforms even LoRA, confirming that hot experts carry the task-relevant signal, and model-level selection lags behind layer-level selection because the global load-balancing loss smooths the per-layer histogram.
 
\begin{table}[t]
\centering
\caption{Expert selection ablation on GSM8K (OlMoE). layer-level hot selection consistently outperforms alternatives. Random selection slightly exceeds LoRA, suggesting that scope restriction itself provides regularization; activation-aware placement amplifies the effect.}
\label{tab:expert_selection}
\small
\begin{tabular}{@{}lcc@{}}
\toprule
\textbf{Selection Strategy} & \textbf{Params} & \textbf{GSM8K Acc.} \\
\midrule
LoRA (all experts) & 4.300\% & 26.37 \\
\midrule
Random ($k=8$, avg. 5 seeds) & 0.678\% & 27.12 \\
Model-level hot & 0.678\% & 28.51 \\
Cold experts & 0.678\% & 25.88 \\
\textbf{layer-level hot (\method{})} & 0.678\% & \textbf{29.49} \\
\bottomrule
\end{tabular}
\end{table}

\subsubsection{Layer Selection}
\label{sec:layer_selection}
 
Table~\ref{tab:layer_selection} ablates which model components receive adapters. Removing attention adapters causes the largest accuracy drop. Gate adapters provide a smaller but consistent benefit. We recommend the full configuration (attention + gate + hot experts) as the default.
 
\begin{table}[t]
\centering
\caption{Layer selection ablation on OlMoE. ``Pure'' = hot experts only; ``+Gate'' adds gating adapters; ``All'' adds attention adapters (our default). Attention adapters provide the largest benefit.}
\label{tab:layer_selection}
\small
\begin{tabular}{@{}lcccc@{}}
\toprule
\textbf{Config} & \textbf{Params} & \textbf{GSM8K} & \textbf{HumanEval} (@1) & \textbf{HEx-PHI} \\
\midrule
\method{} (All) & 0.678\% & \textbf{29.49} & \textbf{17.87} & \textbf{99.06} \\
\method{} (+Gate) & 0.558\% & 28.03 & 17.09 & 97.50 \\
\method{} (Pure) & 0.543\% & 26.46 & 17.07 & 98.43 \\
\bottomrule
\end{tabular}
\end{table}
 
\subsubsection{Number of Hot Experts ($k$)}
\label{sec:k_ablation}
 
Table~\ref{tab:k_ablation} varies $k \in \{2, 4, 8, 16\}$ on GSM8K.
Setting $k = 8$, which matches the number of experts activated per step
in OlMoE, provides the best accuracy-efficiency trade-off, achieving
near-peak accuracy while retaining high throughput. Smaller values of
$k$ save compute but sacrifice accuracy, and larger values offer
diminishing returns.

For other backbones we follow the same principle of matching $k$ to
the routing budget, which yields $k = 2$ for Mixtral. For DeepSeekMoE,
however, the presence of shared experts that are always activated changes
the effective activation distribution among routed experts. Setting $k$
equal to the routed activation count of 6 underperforms, while $k = 18$
also degrades accuracy. A pilot sweep over $k \in \{6, 12, 18\}$
identifies $k = 12$ as the best choice for this architecture. This
suggests that architectures with shared experts may require a moderately
larger $k$ to compensate for the flattened activation distribution
among routed experts.
 
\begin{table}[t]
\centering
\caption{Effect of $k$ (number of hot experts per layer) on GSM8K accuracy and throughput (OlMoE). $k=8$ provides the best trade-off.}
\label{tab:k_ablation}
\small
\begin{tabular}{@{}lccc@{}}
\toprule
$k$ & \textbf{Params (\%)} & \textbf{Accuracy} & \textbf{Throughput (samples/s)} \\
\midrule
2 & 0.27 & 26.72 & 20.99 \\
4 & 0.41 & 26.97 & 20.98 \\
\textbf{8} & \textbf{0.68} & \textbf{29.49} & \textbf{20.08} \\
16 & 1.21 & 29.69 & 16.49 \\
\bottomrule
\end{tabular}
\end{table}
 
\subsubsection{Warm-Up Data Fraction}
\label{sec:warmup_ablation}

Figure~\ref{fig:warmup_validation}(a) shows that the hot experts identified using 10\% of the data largely agree with those identified using the full dataset, and Figure~\ref{fig:warmup_validation}(b) reports the stability of hot-expert identification as the warm-up fraction varies. Using 10\% of the data recovers 87.5\% of full-data hot experts at only 9.2\% training-time overhead. Smaller fractions (1\%) are less reliable, while 50\% provides marginal improvement at $2\times$ the cost. Detailed numerical results are reported in Appendix~\ref{app:warmup_full}.
 
\paragraph{Cross-task preservation.} A natural concern with task-specific 
adapter placement is whether the unadapted cold experts retain their 
pretrained capabilities. We evaluate this in Appendix~\ref{app:crosstask} 
and find that \method{} fine-tuning largely preserves performance on 
non-target tasks, supporting its use in sequential or multi-task 
fine-tuning scenarios.
 
 
 
 
\section{Analysis and Discussion}

\paragraph{Why does adapting cold experts hurt?}
Cold experts are by definition rarely activated for the target task.
When LoRA is attached to a cold expert, its adapter receives gradient
updates from a sparse and unrepresentative subset of tokens. These noisy
updates can perturb the expert's pretrained specialization, which may be
valuable for other capabilities, without contributing meaningfully to the
target task. In effect, adapting cold experts introduces optimization noise
while diluting pretrained knowledge. The layer-level hot-expert selection
in \method{} avoids this by concentrating adaptation capacity where the
model already directs its compute.

\textbf{Variation across backbones.} The accuracy advantage of \method{} 
varies with the activation sparsity of the backbone, with the largest gain on OlMoE (3.12 points), moderate gain on Mixtral (0.55 points), and modest gain on DeepSeekMoE (0.39 points). DeepSeekMoE's shared experts absorb the most dominant activation patterns and are adapted under both methods, leaving less headroom for selective placement. Even in this setting, \method{} matches LoRA's accuracy at roughly four times fewer trainable parameters, indicating that hot-expert selection remains an effective parameter-efficiency strategy across architectures.

\textbf{Connection to structured pruning.} \method{} can be viewed as 
structured pruning of the adapter space, with hot experts as the 
"winning tickets" of adaptation in the sense of the lottery ticket 
hypothesis~\citep{frankle2019lottery}. The random selection baseline 
in Table~\ref{tab:expert_selection} supports this view.

\textbf{Practical guidelines.} For standard MoE architectures, setting 
$k$ equal to the per-step activation count is a reliable starting point. 
Architectures with shared experts require a larger $k$ (roughly twice 
the routed count) because shared experts flatten the routed activation 
distribution. We recommend a 10\% warm-up fraction and adapters on both 
attention and gating components.
 
\section{Limitations}
\method{} has two main limitations. The warm-up stage depends on the 
representativeness of the sampled data. If the warm-up distribution 
diverges from the full training distribution, hot-expert identification 
may be inaccurate. Our experiments show high stability with 10\% of the 
data (Section~\ref{sec:ablations}), but extreme distribution shift could 
degrade selection quality. In addition, \method{} assumes that a small 
subset of experts carries most of the task-relevant signal. For tasks 
requiring broad expert coverage, restricting adaptation to $k$ experts 
per layer may underfit. Adaptive strategies that adjust $k$ per layer 
based on activation entropy are a promising direction for future work.
\section{Broader Impact}

By reducing the cost of adapting MoE language models, \method{} lowers the barrier to fine-tuning capable LLMs for specialized tasks, with positive implications for resource-constrained labs and organizations. However, fine-tuning for harmful applications, including the generation of unsafe, deceptive, or abusive content, also becomes cheaper and more 
accessible. Practitioners should combine \method{} with appropriate safety measures such as content filtering and alignment verification when deploying adapted models. Our safety alignment experiments show that \method{} is compatible with safety fine-tuning, which may partially mitigate this risk.

\section{Conclusion}

We introduced \method{}, a parameter-efficient fine-tuning method for MoE
models that places LoRA adapters only on the most frequently activated
experts at each layer. This simple design simultaneously reduces trainable
parameters by roughly six times relative to LoRA, lowers adapter-induced
FLOPs by 33 to 39\%, increases training throughput by approximately
1.9 times, and improves downstream accuracy, a combination that prior
methods do not achieve. Across three MoE backbones and three task families,
\method{} consistently outperforms LoRA, DoRA, and LoRI variants.
Composing \method{} with LoRI yields \methodri{}, which matches LoRA
accuracy while training only 0.7\% of its trainable parameters. These
results establish activation-aware adapter placement as a practical and
effective principle for scaling PEFT to large MoE language models.
 
\bibliographystyle{plainnat}

\appendix
 
\section{Hyperparameters}
\label{app:hyperparams}

 Appendix A reports the hyperparameters used in all experiments. We use the same LoRA rank and scaling factor for LoRA, DoRA, LoRI, HELLoRA, and HELLoRI variants unless otherwise specified, so that the comparison focuses on adapter placement rather than rank tuning. For full fine-tuning, we use a smaller learning rate and a larger batch size because all model parameters are updated. For PEFT methods, only adapter parameters are optimized, so a larger learning rate is used following common LoRA practice. The warm-up fraction is set to 10\% by default, which provides a favorable trade-off between hot-expert recovery and profiling overhead as analyzed in Appendix E.

The default number of hot experts depends on the routing structure of the backbone. We use $k=8$ for OlMoE, matching its top-8 routing, and $k=2$ for Mixtral, matching its top-2 routing. For DeepSeekMoE, which contains shared experts and routed experts, we use $k=12$ based on the pilot sweep discussed in Section~4.4.3. All reported main results are averaged over 10 random seeds.

\begin{table}[h]
\centering
\caption{Hyperparameters used across all experiments.}
\small
\begin{tabular}{@{}lc@{}}
\toprule
\textbf{Name} & \textbf{Value} \\
\midrule
Adapter rank ($r$) & 32 \\
Batch size & 256 (FFT) / 128 (others) \\
Math learning rate & 4e-5 (FFT) / 4e-4 (others) \\
Code learning rate & 2e-4 (FFT) / 2e-3 (others) \\
Safety learning rate & 2e-5 (FFT) / 4e-4 (others) \\
Math training epochs & 3 \\
Code training epochs & 2 \\
Safety training epochs & 1 \\
Warm-up data fraction & 10\% \\
Hot experts ($k$) & 8 for OlMoE, 2 for Mixtral, 12 for DeepSeekMoE \\
Scaling factor ($\alpha$) & 64 \\
Random seeds & 10 (report average) \\
\bottomrule
\end{tabular}
\end{table}
 
\section{Mixtral-8$\times$7B GSM8K Results}
\label{app:mixtral}
Appendix B provides additional Mixtral-8$\times$7B results for the GSM8K math reasoning setting. Mixtral uses top-2 routing with 8 experts per MoE layer, so HELLoRA selects the top-2 hot experts in each layer after the warm-up profiling stage. This setting tests whether hot-expert placement remains useful when the number of experts per layer is much smaller than in OlMoE.

The results show that HELLoRA improves over standard LoRA while using substantially fewer trainable parameters. HELLoRA reaches 68.28 GSM8K accuracy with 0.31\% trainable parameters, compared with 67.73 accuracy and 1.03\% trainable parameters for LoRA. HELLoRI variants further reduce the parameter budget. In particular, HELLoRI-S uses only 0.02\% trainable parameters while remaining competitive with LoRA, showing that hot-expert placement can also support extreme parameter-constrained adaptation on Mixtral.
 
\begin{table}[h]
\centering
\caption{Full GSM8K results on Mixtral-8$\times$7B ($k=2$, matching top-2 routing).}
\small
\begin{tabular}{@{}lcccccc@{}}
\toprule
& LoRA & LoRI-D & LoRI-S & \method{} & \methodri{}-D & \methodri{}-S \\
\midrule
Params & 1.03\% & 0.59\% & 0.06\% & 0.31\% & 0.17\% & 0.02\% \\
Accuracy & 67.73 & 64.84 & 63.98 & \textbf{68.28} & 64.61 & 66.17 \\
\bottomrule
\end{tabular}
\end{table}
 
\section{Layer-level Expert Activation Patterns}
\label{app:activation}

We visualize expert activation distributions for all three backbones
to complement the motivation presented in Section~1.

\begin{figure}[h]
    \centering
    \includegraphics[width=\textwidth]{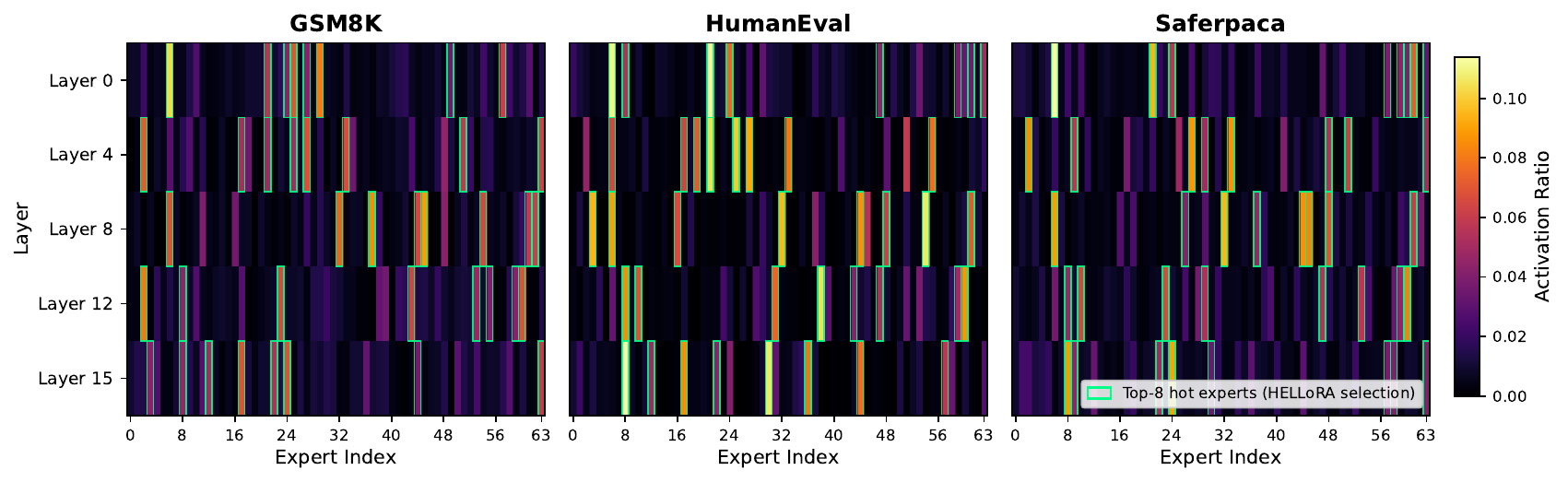}
    \caption{\textbf{Expert activation heatmaps on OlMoE-1B-7B.}
    Each panel shows one task. Rows correspond to MoE layers
    and columns to expert indices (0 to 63).
    Color intensity indicates activation frequency.
    Green boxes mark the top-8 hot experts per layer.
    Hot experts vary across both layers and tasks.}
    \label{fig:olmoe_heatmap}
\end{figure}

For OlMoE (Figure~\ref{fig:olmoe_heatmap}), we report activation
heatmaps across all three downstream tasks (GSM8K, HumanEval, Saferpaca)
at layers 0, 4, 8, 12, and 15. Two consistent phenomena emerge.
First, within each layer, activation is highly skewed toward
a small number of hot experts. Second, the identity of hot experts
varies across-tasks, confirming that expert importance is both
layer-specific and task-specific.

\begin{figure}[h]
    \centering
    \includegraphics[width=0.55\textwidth]{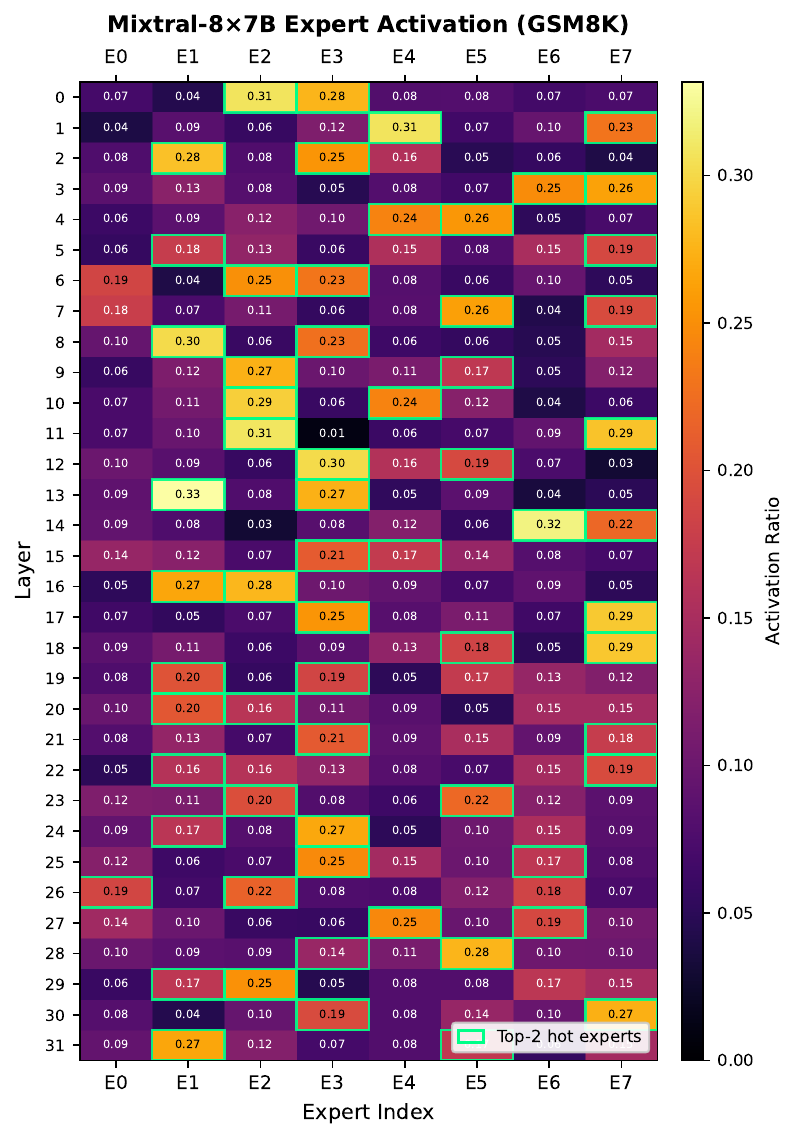}
    \caption{\textbf{Expert activation heatmap for Mixtral-8$\times$7B on GSM8K.}
    Each cell shows the activation ratio of one expert at one layer.
    Green boxes mark the top-2 hot experts per layer.
    Despite having only 8 experts with top-2 routing,
    activation remains highly skewed across all 32 layers,
    with the two most active experts typically
    accounting for over 50\% of all token assignments.}
    \label{fig:mixtral_activation}
\end{figure}

For Mixtral-8$\times$7B (Figure~\ref{fig:mixtral_activation}),
we report activation ratios on GSM8K across all 32 MoE layers.
Despite having only 8 experts per layer with top-2 routing,
activation remains strongly skewed, with the two most active
experts typically accounting for over 50\% of token assignments
at each layer. The identity of hot experts shifts across layers,
indicating that the layer-specific activation pattern observed
in OlMoE generalizes to architectures with fewer but more
heavily loaded experts.

\begin{figure}[h]
    \centering
    \includegraphics[width=0.85\textwidth]{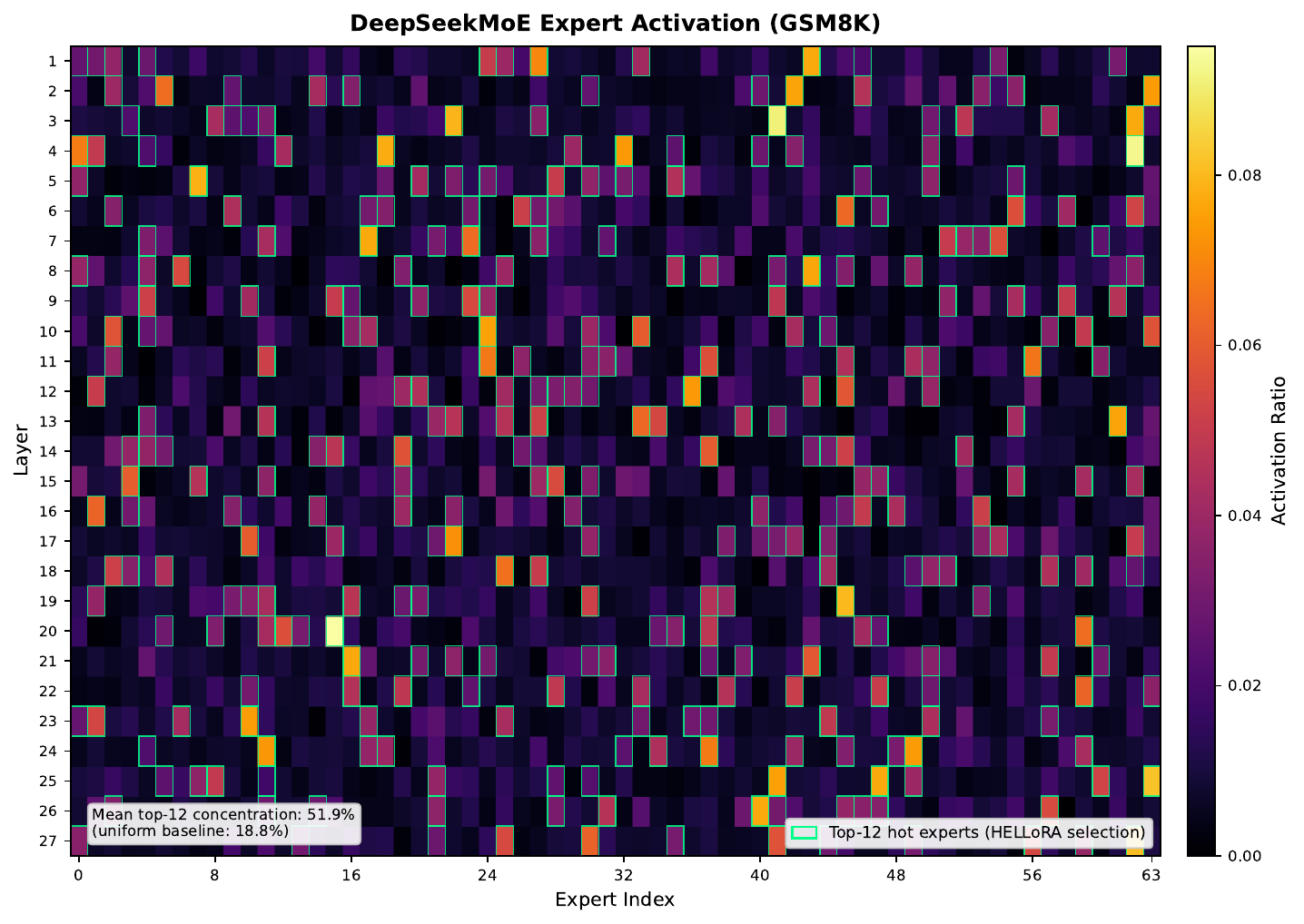}
    \caption{\textbf{Expert activation heatmap for DeepSeekMoE on GSM8K.}
    Rows correspond to 27 MoE layers and columns to 64 routed expert indices.
    Green boxes mark the top-12 hot experts per layer.
    The top-12 experts account for 51.9\% of activations on average,
    compared with 18.8\% under uniform allocation.
    The identity of hot experts shifts across layers,
    confirming that activation skew is a general property
    of MoE architectures.}
    \label{fig:deepseek_activation}
\end{figure}

For DeepSeekMoE (Figure~\ref{fig:deepseek_activation}),
we report activation ratios on GSM8K across all 27 MoE layers,
each containing 64 routed experts with top-6 routing.
The top-12 experts per layer account for 51.9\% of activations on average,
far exceeding the 18.8\% expected under uniform allocation.
As with OlMoE and Mixtral, the identity of hot experts varies
substantially across layers, and no single expert is consistently
dominant. This confirms that the activation skew exploited
by \method{} is a general property of MoE architectures,
independent of the number of experts, routing strategy,
or the presence of shared experts.

\section{Throughput Comparison with Layer Selection Variants}
\label{app:throughput_layer}

In Section~\ref{sec:layer_selection} we reported the accuracy impact of different layer selection configurations. Here we complement that analysis by reporting the corresponding training throughput on OlMoE GSM8K. Table~\ref{tab:layer_throughput} shows the results under three 
configurations. "Pure" attaches adapters only to expert FFNs, "+Gate" additionally attaches adapters to the gating networks, and "All" further adds adapters to the attention projections (our default).

\begin{table}[h]
\centering
\caption{Throughput (samples/s) across layer selection variants on 
OlMoE GSM8K. \method{} and its variants maintain approximately 
$1.8$ to $1.9\times$ the throughput of LoRA and LoRI variants across 
all three configurations.}
\label{tab:layer_throughput}
\small
\begin{tabular}{@{}lccc@{}}
\toprule
\textbf{Method} & \textbf{Pure} & \textbf{+Gate} & \textbf{All} \\
\midrule
LoRA          & 10.01 & 11.97 & 10.44 \\
LoRI-D        & 10.95 & 10.98 & 10.65 \\
LoRI-S        & 10.77 & 10.91 & 10.00 \\
\method{}     & 19.60 & 19.20 & 19.08 \\
\methodri{}-D & 18.41 & 19.03 & 19.11 \\
\methodri{}-S & 19.45 & 19.02 & 19.07 \\
\bottomrule
\end{tabular}
\end{table}

Two observations emerge. First, \method{} maintains its throughput 
advantage over LoRA and LoRI variants across all three configurations, 
indicating that the speedup comes from the hot-expert placement strategy 
itself rather than from any particular choice of which non-expert 
modules receive adapters. Second, adding attention adapters incurs 
only a small throughput cost for \method{} (from 19.60 to 19.08, about 
2.7\%) while providing substantial accuracy gains 
(Table~\ref{tab:layer_selection}). This supports our recommendation to 
use the full configuration (attention, gating, and hot experts) as the 
default setting.

\section{Warm-Up Data Fraction Details}
\label{app:warmup_full}

Section~\ref{sec:warmup_ablation} discussed the trade-off between 
warm-up data fraction and hot-expert recovery. Table~\ref{tab:warmup_full} 
provides the precise numerical values underlying Figure~\ref{fig:warmup_validation}(b). 
We measure two stability metrics. The Jaccard overlap quantifies the 
similarity between the hot-expert sets identified using a partial 
warm-up and those identified using the full dataset. The coverage rate 
reports the percentage of full-data hot experts that are included in 
the partial-data selection.

\begin{table}[h]
\centering
\caption{Hot-expert recovery and wall-clock overhead as a function of 
warm-up fraction (OlMoE, GSM8K). Overlap is measured against the 
hot-expert sets identified using 100\% of the training data. The 10\% 
default setting offers the best balance between selection quality and 
computational cost.}
\label{tab:warmup_full}
\begin{tabular}{lcccc}
\toprule
Fraction & Jaccard & Coverage & Time (s) & Overhead \\
\midrule
1\%   & 0.753 & 85.9\% & 167.9 & 7.6\%  \\
5\%   & 0.753 & 85.9\% & 175.3 & 8.0\%  \\
10\%  & 0.778 & 87.5\% & 201.9 & 9.2\%  \\
20\%  & 0.778 & 87.5\% & 268.2 & 12.2\% \\
50\%  & 0.803 & 89.1\% & 455.5 & 20.7\% \\
\bottomrule
\end{tabular}
\end{table}

Two observations support the choice of 10\% as the default. First, the 
coverage rate plateaus around 87 to 89\% across the 10 to 50\% range, 
indicating that further increasing the warm-up fraction yields 
diminishing returns. Second, the overhead grows roughly linearly with 
the warm-up fraction. The 10\% setting captures most of the achievable 
selection quality at the lowest cost beyond which additional data 
provides limited benefit.

\section{Analysis of LoRAMoE on MoE Backbones}
\label{app:loramoe_rank32}

LoRAMoE~\citep{dou2024loramoe} was originally designed for dense 
backbones. It inserts $N$ LoRA-based experts with a learned router into 
each FFN layer to create an MoE-style adapter structure. When applied 
to OlMoE, whose FFN layers already perform top-$k$ routing over 64 
experts, LoRAMoE effectively introduces a second layer of routing on 
top of the native one. This appendix documents our LoRAMoE experiments 
and analyzes why this configuration underperforms \method{}.

\paragraph{Configurations.} We follow the original LoRAMoE paper with 
rank $r = 4$ and $N = 6$ LoRA experts per FFN, using a learned gating 
network over the experts. For a capacity-matched comparison with 
\method{} (which uses $r = 32$), we also evaluate LoRAMoE with $r = 32$. 
Both configurations use the same training data (GSM8K train split), 
optimizer, and seeds as other baselines in Table~\ref{tab:main_olmoe}.

\paragraph{Results.} Table~\ref{tab:loramoe_rank32} summarizes GSM8K 
accuracy and trainable-parameter cost. Increasing LoRAMoE's rank from 
4 to 32 raises trainable parameters from 3.59\% to 21.03\% but reduces 
accuracy from 26.88 to 20.70. In contrast, \method{} at $r = 32$ achieves 
29.49 accuracy with only 0.70\% trainable parameters.

\begin{table}[h]
\centering
\caption{LoRAMoE on OlMoE GSM8K. Adding adapter capacity to LoRAMoE 
does not translate to higher accuracy. \method{} achieves both higher 
accuracy and substantially fewer trainable parameters.}
\label{tab:loramoe_rank32}
\begin{tabular}{lccc}
\toprule
Method & Rank & Params & GSM8K \\
\midrule
LoRAMoE   & 4  & 3.59\%  & 26.88 \\
LoRAMoE   & 32 & 21.03\% & 20.70 \\
\method{} & 32 & 0.70\%  & \textbf{29.49} \\
\bottomrule
\end{tabular}
\end{table}

\paragraph{Interpretation.} We interpret these results as evidence of 
a structural mismatch. On a dense backbone, LoRAMoE's adapter-level 
routing introduces MoE-style specialization that did not previously 
exist. On an MoE backbone, this specialization is redundant with the 
native router, and the two routing layers must co-adapt through training. 
Increasing adapter rank amplifies the interference between the two 
routers rather than resolving it, which leads to degraded accuracy at 
higher capacity. \method{} avoids this issue by operating at the placement 
level. It selects which native experts receive adapters and leaves the 
backbone's routing intact. This design aligns adaptation with the 
backbone's existing sparse structure rather than adding a competing 
structure on top.

\section{Cross-Task Preservation Analysis}
\label{app:crosstask}

A natural concern with task-specific adapter placement is whether 
\method{} fine-tuning on one task degrades performance on other tasks 
that may rely on the unadapted cold experts. To investigate this, we 
measure cross-task performance before and after \method{} fine-tuning. 
For each target task, we fine-tune \method{} on that task's training 
data and then evaluate the resulting model on all three downstream 
tasks. Table~\ref{tab:crosstask} reports the results.

\begin{table}[h]
\centering
\caption{Cross-task evaluation on OlMoE. Each row corresponds to 
\method{} fine-tuned on one target task, evaluated across all three 
tasks. B denotes the score before fine-tuning (the pretrained baseline) 
and A denotes the score after fine-tuning. Cells with dashes indicate 
the target task itself, whose post-finetuning score is reported in 
Table~\ref{tab:main_olmoe}.}
\label{tab:crosstask}
\begin{tabular}{lcccccc}
\toprule
Target & GSM8K\textsubscript{B} & GSM8K\textsubscript{A} & HumanEval\textsubscript{B} & HumanEval\textsubscript{A} & SAFE\textsubscript{B} & SAFE\textsubscript{A} \\
\midrule
GSM8K     & --   & --   & 13.41 & 13.78 & 66.75 & 58.43 \\
HumanEval & 1.66 & 5.86 & --    & --    & 66.75 & 63.01 \\
SAFE      & 1.66 & 2.83 & 13.41 & 15.18 & --    & --    \\
\bottomrule
\end{tabular}
\end{table}

Two patterns emerge from the table. First, on non-target tasks, 
\method{} fine-tuning generally preserves or even improves performance. 
For example, fine-tuning on GSM8K slightly increases HumanEval from 
13.41 to 13.78, and fine-tuning on HumanEval improves GSM8K from 1.66 
to 5.86. The improvements suggest that adapting hot experts on one task 
does not destructively interfere with other capabilities and may even 
transfer beneficially. Second, the only notable degradation appears on 
the safety benchmark after GSM8K fine-tuning (66.75 to 58.43). This 
modest drop is consistent with prior findings that math fine-tuning 
can mildly affect safety alignment~\citep{qi2024finetuning}, and is 
not specific to \method{}.

These results support the interpretation that freezing cold experts 
acts as a structured regularizer that protects pretrained capabilities. 
This property makes \method{} a natural candidate for sequential or 
multi-task fine-tuning scenarios where capability preservation is 
critical.

\section{Cross-Backbone Results across All Methods and Tasks}
\label{app:cross_backbone_full}

Table~\ref{tab:cross_backbone} in the main text presents a summary of 
\method{} versus LoRA on three tasks across three backbones. Here we 
provide the complete results, including LoRI and \methodri{} variants, 
on Mixtral-8$\times$7B (Table~\ref{tab:mixtral_full}) and DeepSeekMoE 
(Table~\ref{tab:deepseek_full}). The OlMoE results are reported in 
Table~\ref{tab:main_olmoe} of the main text.

\begin{table}[h]
\centering
\caption{Full results on Mixtral-8$\times$7B across three tasks. 
\method{} achieves the best performance on math reasoning and safety 
alignment, while \methodri{}-D leads on code generation under an 
extreme parameter budget. Results are mean values over 10 random seeds.}
\label{tab:mixtral_full}
\small
\begin{tabular}{lccccc}
\toprule
Method & Params & GSM8K & HumanEval (@1 / @5 / @10) & HEx-PHI \\
\midrule
LoRA          & 1.03\% & 67.73 & 39.99 / 49.47 / 52.72 & 93.94 \\
LoRI-D        & 0.59\% & 64.84 & 44.72 / 53.81 / 56.67 & 93.33 \\
LoRI-S        & 0.06\% & 63.98 & 43.11 / 52.87 / 56.88 & 93.64 \\
\midrule
\method{}     & 0.31\% & \textbf{68.28} & 44.89 / 53.97 / 57.41 & \textbf{96.67} \\
\methodri{}-D & 0.17\% & 64.61 & \textbf{47.41} / \textbf{57.53} / \textbf{60.43} & 94.85 \\
\methodri{}-S & 0.02\% & 66.17 & 43.08 / 54.03 / 58.25 & 96.06 \\
\bottomrule
\end{tabular}
\end{table}

\begin{table}[h]
\centering
\caption{Full results on DeepSeekMoE across three tasks. \method{} 
achieves the best performance on every metric while using significantly 
fewer trainable parameters than LoRA. Results are mean values over 10 
random seeds.}
\label{tab:deepseek_full}
\small
\begin{tabular}{lccccc}
\toprule
Method & Params & GSM8K & HumanEval (@1 / @5 / @10) & HEx-PHI \\
\midrule
LoRA          & 3.545\% & 35.15 & 27.92 / 34.98 / 38.52 & 97.87 \\
LoRI-D        & 1.671\% & 30.63 & 26.77 / 33.48 / 37.53 & 97.48 \\
LoRI-S        & 0.167\% & 31.87 & 27.68 / 34.72 / 38.22 & 96.56 \\
\midrule
\method{}     & 0.823\% & \textbf{35.54} & \textbf{31.15} / \textbf{36.65} / \textbf{39.19} & \textbf{98.49} \\
\methodri{}-D & 0.395\% & 33.59 & 29.27 / 35.78 / 37.61 & 98.13 \\
\methodri{}-S & 0.040\% & 35.23 & 26.62 / 32.19 / 34.63 & 94.68 \\
\bottomrule
\end{tabular}
\end{table}

The results show that the patterns observed on OlMoE consistently 
transfer to Mixtral and DeepSeekMoE. \method{} achieves the best 
performance on math reasoning and safety alignment across all three 
backbones, and outperforms every other baseline on code generation 
except in one configuration. The exception occurs on Mixtral, where 
\methodri{}-D outperforms \method{} on HumanEval at less than half 
the trainable parameter budget. We attribute this to the extreme 
sparsity of Mixtral's top-2 routing, which amplifies the regularization 
benefits of additional weight masking. Even in this case, \method{} 
remains substantially above all other baselines. Overall, hot-expert 
placement consistently outperforms standard PEFT methods, while 
\methodri{} provides a complementary option for scenarios with severe 
parameter constraints.
 






\end{document}